\documentclass[letterpaper, 10 pt, conference]{ieeeconf}  % Comment this line out if you need a4paper

\IEEEoverridecommandlockouts                              % This command is only needed if you want to use the \thanks command

\usepackage{graphics} % for pdf, bitmapped graphics files
\usepackage{epsfig} % for postscript graphics files
\usepackage{mathptmx} % assumes new font selection scheme installed
\usepackage{times} % assumes new font selection scheme installed
\usepackage{amsmath} % assumes amsmath package installed
\usepackage{amssymb}  % assumes amsmath package installed
\usepackage{bm}
\usepackage{multirow}
\usepackage{graphicx}
\usepackage[final]{changes}
\usepackage{optidef}
\usepackage{array}
\usepackage{colortbl}
\DeclareGraphicsExtensions{.pdf,png}

\title{\LARGE \bf
Legged Robot State Estimation With Invariant Extended Kalman Filter Using Neural Measurement Network
}

\author{Donghoon Youm$^{1}$, Hyunsik Oh$^{1}$, Suyoung Choi$^{1}$, Hyeongjun Kim$^{1}$, Jemin Hwangbo$^{1}$
\thanks{$^{1}$Korea Advanced Institute of Science and Technology, Yuseong-gu,
        Daejeon, 34141, Republic of Korea
        {\tt\small {ydh0725, ohsik1008, swimchoy, kaist0914, jhwangbo}@kaist.ac.kr}}%
}

\begin{document}
\bstctlcite{IEEEexample:BSTcontrol}
\maketitle
\thispagestyle{empty}
\pagestyle{empty}

%%%%%%%%%%%%%%%%%%%%%%%%%%%%%%%%%%%%%%%%%%%%%%%%%%%%%%%%%%%%%%%%%%%%%%%%%%%%%%%%
% \the\columnwidth

%% editing comment

%\newcommand{\cmt}[1]{\textcolor{red}{\textbf {#1}}}
\newcommand{\cmt}[1]{}

\newcommand{\updated}[1]{\textcolor{blue}{{#1}}}
\newcommand{\hyunyoung}[1]{\textcolor{orange}{{Hyunyoung: #1}}}
\newcommand{\donghoon}[1]{\textcolor{cyan}{{Donghoon: #1}}} 
\newcommand{\sehoon}[1]{\textcolor{red}{{Sehoon: #1}}}

\newcommand{\newtext}[1]{#1}
\newcommand{\eqnref}[1]{Equation~(\ref{eq:#1})}
\newcommand{\figref}[1]{Figure~\ref{fig:#1}}
\newcommand{\tabref}[1]{Table~\ref{tab:#1}}
\newcommand{\secref}[1]{Section~\ref{sec:#1}}

%% ignore text
\long\def\ignorethis#1{}

%% abbreviations
\newcommand{\etal}{{\em{et~al.}\ }}
\newcommand{\eg}{e.g.\ }
\newcommand{\ie}{i.e.\ }

%% reference shortcuts
\newcommand{\figtodo}[1]{\framebox[0.8\columnwidth]{\rule{0pt}{1in}#1}}

%\renewcommand{\eqref}[1]{Equation~(\ref{eq:#1})}

%% frequently used mathematical structures

\newcommand{\pdd}[3]{\ensuremath{\frac{\partial^2{#1}}{\partial{#2}\,\partial{#3}}}}

%% New commands for Sehoon!
\newcommand{\mat}[1]{\ensuremath{\mathbf{#1}}}
\newcommand{\set}[1]{\ensuremath{\mathcal{#1}}}

% math macros
\newcommand{\vc}[1]{\ensuremath{\mathbf{#1}}}
\newcommand{\vEndEff}{\ensuremath{\vc{d}}}
\newcommand{\vRelMove}{\ensuremath{\vc{r}}}
\newcommand{\sSet}{\ensuremath{S}}

\newcommand{\vControl}{\ensuremath{\vc{u}}}
\newcommand{\vPoint}{\ensuremath{\vc{p}}}
\newcommand{\sSpringCoef}{{\ensuremath{k_{s}}}}
\newcommand{\sDamperCoef}{{\ensuremath{k_{d}}}}
\newcommand{\vHandle}{\ensuremath{\vc{h}}}
\newcommand{\vForce}{\ensuremath{\vc{f}}}

\newcommand{\mTransChain}{\ensuremath{\vc{W}}}
\newcommand{\mRotateTrans}{\ensuremath{\vc{R}}}
\newcommand{\sJoint}{\ensuremath{q}}
\newcommand{\vJoint}{\ensuremath{\vc{q}}}
\newcommand{\mJoint}{\ensuremath{\vc{Q}}}
\newcommand{\mMass}{\ensuremath{\vc{M}}}
\newcommand{\sMass}{\ensuremath{{m}}}
\newcommand{\vGravity}{\ensuremath{\vc{g}}}
\newcommand{\vConstr}{\ensuremath{\vc{C}}}
\newcommand{\sConstr}{\ensuremath{C}}
\newcommand{\vCOM}{\ensuremath{\vc{x}}}
\newcommand{\sGeneralForce}[1]{\ensuremath{Q_{#1}}}
\newcommand{\vStateVar}{\ensuremath{\vc{y}}}
\newcommand{\vControlVar}{\ensuremath{\vc{u}}}
\newcommand{\tr}[1]{\ensuremath{\mathrm{tr}\left(#1\right)}}

%%%%%%%%%%%%%%%%%%%%%%%%%%%%%%%%%%%%%%%%%%%%%%%%%%%%%%%%%%%%%%%%%%%
%
% Here are a bunch of macros, mostly for math.
%
%%%%%%%%%%%%%%%%%%%%%%%%%%%%%%%%%%%%%%%%%%%%%%%%%%%%%%%%%%%%%%%%%%%

\renewcommand{\choose}[2]{\ensuremath{\left(\begin{array}{c} #1 \\ #2 \end{array} \right )}}

\newcommand{\gauss}[3]{\ensuremath{\mathcal{N}(#1 | #2 ; #3)}}

\newcommand{\pctab}{\hspace{0.2in}}
\newenvironment{pseudocode} {\begin{center} \begin{minipage}{\textwidth}
                             \normalsize \vspace{-2\baselineskip} \begin{tabbing}
                             \pctab \= \pctab \= \pctab \= \pctab \=
                             \pctab \= \pctab \= \pctab \= \pctab \= \\}
                            {\end{tabbing} \vspace{-2\baselineskip}
                             \end{minipage} \end{center}}
\newenvironment{items}      {\begin{list}{$\bullet$}
                              {\setlength{\partopsep}{\parskip}
                                \setlength{\parsep}{\parskip}
                                \setlength{\topsep}{0pt}
                                \setlength{\itemsep}{0pt}
                                \settowidth{\labelwidth}{$\bullet$}
                                \setlength{\labelsep}{1ex}
                                \setlength{\leftmargin}{\labelwidth}
                                \addtolength{\leftmargin}{\labelsep}
                                }
                              }
                            {\end{list}}
\newcommand{\newfun}[3]{\noindent\vspace{0pt}\fbox{\begin{minipage}{3.3truein}\vspace{#1}~ {#3}~\vspace{12pt}\end{minipage}}\vspace{#2}}

\newcommand{\key}{\textbf}
\newcommand{\fun}{\textsc}
%\def\shortcite{\def\citename##1{}\@internalcite}

% Local Variables:
% TeX-master: "paper"
% End:

\begin{abstract}

This paper introduces a novel proprioceptive state estimator for legged robots that combines model-based filters and deep neural networks. Recent studies have shown that neural networks such as multi-layer perceptron or recurrent neural networks can estimate the robot states, including contact probability and linear velocity. Inspired by this, we develop a state estimation framework that integrates a neural measurement network (NMN) with an invariant extended Kalman filter. We show that our framework improves estimation performance in various terrains. Existing studies that combine model-based filters and learning-based approaches typically use real-world data. However, our approach relies solely on simulation data, as it allows us to easily obtain extensive data. This difference leads to a gap between the learning and the inference domain, commonly referred to as a sim-to-real gap. We address this challenge by adapting existing learning techniques and regularization. To validate our proposed method, we conduct experiments using a quadruped robot on four types of terrain: \textit{flat}, \textit{debris}, \textit{soft}, and \textit{slippery}. We observe that our approach significantly reduces position drift compared to the existing model-based state estimator.

\end{abstract}

\section{introduction}
% Recent breakthroughs in legged robots have demonstrated notable research outcomes~\cite{hong2022agile, choi2023learning, kim2023not}. Accurate and reliable state estimation underlies these advances, emerging as a crucial stepping stone for achieving dynamic control, regardless of the chosen control methodology.
% In the realm of model-based control, states such as orientation and linear velocity are explicitly incorporated in predicting future trajectory, and the estimation accuracy directly affects the control performance~\cite{ding2021representation, grandia2023perceptive}. Learning-based control either explicitly estimates states or employs a neural network for implicit estimation~\cite{siekmann2021blind, ji2022concurrent}. A study by Yu et al.~\cite{yu2023identifying} revealed the relative importance of each state for control, as estimated in this way, through saliency analysis.
Recent breakthroughs in dynamic locomotion of legged robots have demonstrated the ability to navigate various terrains such as steel structures~\cite{hong2022agile}, mountainous terrains~\cite{kim2023not}, and sandy fields~\cite{choi2023learning}. 
These remarkable developments have stemmed from the advancements in various control methods. One of these control methodologies is model predictive control, 
% One of the control methodologies for dynamic locomotion is model predictive control, 
which plans the future trajectories based on the current robot state. Therefore, the accurate estimation of the current robot states directly affects control performance. Another control methodology, learning-based control, estimates the robot states explicitly or implicitly through a neural network and then uses it for control~\cite{siekmann2021blind, ji2022concurrent}. Regardless of the control methodology, accurate and reliable state estimation is a crucial stepping stone for dynamic control.

State estimation using proprioceptive sensors, including magnetic encoders and Inertial Measurement Unit~(IMU), is a low-cost, lightweight, and power-efficient solution. These sensors provide direct access to joint angles, joint velocity, acceleration, and angular velocity. Processing these proprioceptive data with a state estimation algorithm can reconstruct the unknown states such as position, orientation, and linear velocity. 

% Adding exteroceptive sensors such as Lidar and cameras may further enhance the accuracy of the estimates. However, their vulnerability to lighting variations, low-texture environments, and motion blur often drives the estimation inconsistently. Furthermore, they are costly, heavy, and power-intensive, which makes the use of exteroceptive sensors less practical. Research on proprioceptive sensor-based estimators for legged robots has been actively pursued for these reasons.
Researchers have shown that the state estimation accuracy can be further improved by adding exteroceptive sensors such as Lidar and cameras~\cite{wisth2022vilens}. However, they are vulnerable to lighting variations, low-texture environments, and motion blur. Consequently, they may not be a suitable estimator for dynamic control in which state estimation results are important. Research on proprioceptive-sensor-based estimators for legged robots has been actively pursued for these reasons.

\begin{figure}[tp]
\centering
\includegraphics[width=\columnwidth]{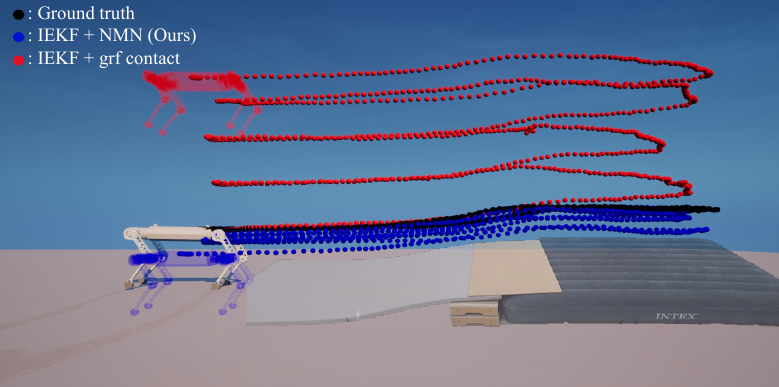}
\caption{By combining our Neural Measurement Network (NMN) with a model-based filter, our method can reduce the position estimation error for the entire trajectory of the Raibo robot on an air mattress using only proprioceptive sensors by one-third compared to the purely model-based methods.}
\label{fig:teaser}
% \vspace{-0.4cm}
\end{figure}

Bloesch et al.~\cite{bloesch2012state} introduced an EKF, which uses IMU and contact motion models. This approach incorporates contact foot kinematics as a measurement, assuming non-slip contact with the feet. The following research has employed the Unscented Kalman Filter (UKF) or factor graph methods to improve estimation performance~\cite{bloesch2013state, hartley2018legged}. A contact-aid Invariant Extended Kalman Filter (IEKF), applied by Hartley et al.~\cite{hartley2020contact} to legged robots, achieves enhanced convergence and robustness by utilizing the properties of the filter`s state space formulated as a Lie group. Research has indicated that the IEKF offers superior convergence and consistency compared to the conventional quaternion-based extended Kalman filter (QEKF)~\cite{sola2017quaternion}.

% Unfortunately, all these approaches are fragile to unmodeled effects such as foot slippage or ambiguous contact in soft terrain, even when using contact sensors, due to their underlying non-slip assumption. To address these issues, Bloesch et al.~\cite{bloesch2013state} devised a stochastic filtering method, handling outlier measurements through the thresholding of the Mahalanobis distance of innovation. Kim et al.~\cite{kim2021legged} introduced a slip rejection method to enhance the robustness to foot slip of estimators, which used kinematic measurements for legged robots. They identified slip via the contact foot's velocity and modified the kinematics model's covariance, improving robustness against foot slips.
Due to their underlying non-slip assumption, the aforementioned approaches are fragile to un-modeled situations such as foot slip or ambiguous contact in soft terrain, even when using contact sensors. To address these issues, Bloesch et al.~\cite{bloesch2013state} devised a stochastic filtering method, handling outlier measurements through the threshold of the Mahalanobis distance of innovation. Kim et al.~\cite{kim2021legged} introduced a slip rejection method that identifies slip via the contact foot's velocity and modified the kinematics model's covariance, improving robustness against foot slip.

% The data-driven method is another promising approach for state estimation. With this approach, researchers have transformed the inertial odometry into a regression problem, and they manipulate or denoise raw sensor measurements to estimate orientation or position~\cite{chen2018ionet, herath2020ronin, brossard2020denoising, zhang2021imu, esfahani2019aboldeepio, li2021inertial}. Compared to traditional methods, these studies have demonstrated impressive accuracy in datasets for Micro Aerial Vehicle (MAV) and pedestrians. All these studies have been conducted using real-world datasets. Unfortunately, both the data collection and postprocessing are extremely expensive and laborious, especially for legged robots, which can be damaged when controlled immaturely. In this light, research has been extensively conducted to train control policies only with simulated data. However, methods for training state estimators with simulations have not been rigorously studied.
The data-driven method is another promising approach for state estimation. Recent research in inertial navigation systems (INS) and inertial odometry (IO) have leveraged machine learning to estimate the motion directly from IMU data. Pioneering work by Chen et al.~\cite{chen2018ionet} introduced IoNet, which used a two-layer bidirectional Long Short-Term Memory (Bi-LSTM) to predict a pedestrian's location transforms in polar coordinates from raw IMU data. Yan et al.~\cite{herath2020ronin} presented RoNIN, which infers the velocity and heading direction of pedestrians directly from raw IMU data, and they evaluated the performance of various backbone networks such as ResNet, LSTM, and Time Convolutional Network (TCN). In a different direction, Brossard et al.~\cite{brossard2020denoising} and Zhang et al.~\cite{zhang2021imu} improved the estimation performance through learning-based methods to denoise raw IMU data rather than estimating the pose directly. 
% Compared to traditional methods, these studies have demonstrated impressive accuracy in pedestrian, drone, and wheeled-vehicle datasets.

% Numerous researchers have investigated combining the strengths of model-based and learning-based estimation strategies. A prominent trend has been the fusion of the traditional EKF framework with neural networks, applicable to various domains. This includes pedestrian tracking~\cite{liu2020tlio, sun2021idol}, wheeled vehicles~\cite{brossard2020ai, jin2023learning}, quadrotors~\cite{zhang2022dido, cioffi2023learned}, and quadruped robot platforms~\cite{buchanan2022learning, lin2021legged}. 
Many researchers have investigated combining the strengths of both model-based and learning-based estimation strategies. The fusion of the EKF framework and neural networks is applied to various domains including pedestrian tracking~\cite{liu2020tlio, sun2021idol}, wheeled vehicles~\cite{brossard2020ai, jin2023learning}, quadrotors~\cite{zhang2022dido, cioffi2023learned}, and quadruped robots~\cite{buchanan2022learning, lin2021legged}. 
% Buchanan et al.~\cite{buchanan2022learning} learned relative displacement measurements and covariance from buffered IMU data and applied them to a legged robot. They then integrated these measurements with the EKF and factor graph framework using Pronto~\cite{camurri2020pronto}. 
% \added{In similar manner, Liu et al.~\cite{liu2020tlio} applied to pedestrian dataset, and Cioffi et al.~\cite{cioffi2023learned} employed to drone dataset.}
% They outperform existing model-based or learning-based methods, but their reliance on relative position measurements is a limitation. This approach prevents them from benefiting from the convergence and robustness of the IEKF framework, as their state space does not satisfy the Lie Group properties. Lin et al.~\cite{lin2021legged} proposed an alternative approach that used motion capture or Visual-Inertial Odometry (VIO) to learn foot contact states for legged robot systems, and they used this learned information for leg kinematics measurements at IEKF.
Liu et al.~\cite{liu2020tlio} developed a neural network that learns relative displacement measurements and covariance from buffered IMU data, and integrated the neural network with the EKF. 
Buchanan et al.\cite{buchanan2022learning} expanded this work to quadruped robots by incorporating the factor graph framework. 
% Cioffi et al.~\cite{cioffi2023learned} focused solely on learning relative displacement measurements for integration with the EKF.
% \added{In a similar manner, Liu et al.~\cite{liu2020tlio} applied to pedestrian datasets, and Cioffi et al.~\cite{cioffi2023learned} employed to drone datasets.} 
Although these approaches outperform existing model-based or learning-based methods, their reliance on relative displacement measurements restricts compatibility with the IEKF framework. 
This limitation occurs because the state space of the estimator using relative displacement measurements does not satisfy the Lie Group property. 
% To this end, Lin et al.~\cite{lin2021legged} proposed a neural network that learns foot contact states for quadruped robots, and used these states for leg kinematics measurements in the IEKF.
To this end, Lin et al.~\cite{lin2021legged} proposed a learning-based contact estimator for quadruped robots and integrated it with contact-aid IEKF to bypass these limitations and the need for contact sensors.

% Generating data is an essential requirement for training. For this reason, existing studies either relied on well-established real-world datasets~\cite{antonini2018blackbird, geiger2013vision} for training or required efforts to gather data. Unfortunately, both the data collection and postprocessing are extremely expensive and laborious, especially for legged robots, which can be damaged when controlled immaturely. In this light, research has been extensively conducted to train control policies only with simulated data. However, methods for training state estimators to estimate position and orientation with simulations have not been rigorously studied.
Learning-based methods have either relied on well-established real-world datasets~\cite{antonini2018blackbird, geiger2013vision} or required efforts to gather data, as a sufficient amount of data is essential for training. Unfortunately, the data collection and post-processing are extremely expensive and laborious, especially for legged robots. In this light, the existing research in legged robotics has been focusing on using only simulation data. However, methods for training a state estimator to estimate position and orientation relying solely on simulation data have not been rigorously studied.

% \deleted{Inspired by Ji et al.{~\cite{ji2022concurrent}}, which showed a learning-based control framework training state estimator concurrently through a supervised approach, }We present a novel proprioceptive state estimator that integrates a neural measurement network (NMN) with an IEKF. Our approach capitalizes on expansive simulation datasets to curtail the requisite effort while taking advantage of IEKF. In this study, we choose linear velocity expressed in the body frame and contact probability as a neural measurement and use it as a measurement model in the correction stage of the IEKF. The key challenge to our approach is the sim-to-real gap -- the discrepancy between the simulated environment and the real world. We meticulously analyze the efficacy of existing learning techniques, such as early stopping and removing randomization, and introduce a novel smoothness loss in supervised learning to reduce such inconsistency. Our approach shows reliability superior to a model-based estimator with strict contact assumption under challenging scenarios, including debris, soft ground, and slippery terrains. We summarize our contributions as follows.
We present a novel proprioceptive state estimator that integrates a neural measurement network (NMN) with an IEKF. Our approach capitalizes on extensive simulation datasets to train NMN while making use of the superior convergence and robustness of IEKF. In this study, we choose linear velocity expressed in the body frame and contact probability as an output of NMN and use them as observations in the correction step of the IEKF. The key challenge to our approach is the sim-to-real gap: the discrepancy between the simulation environment and the real world. To reduce this discrepancy, we test existing learning techniques, such as early stopping and domain randomization, in our training and meticulously analyze their efficiency.
% We also introduce a novel smoothness loss in supervised learning for the regularization of NMN output.
We also introduce a smoothness loss in supervised learning for the regularization of NMN output.
Under diverse scenarios such as flat, debris, soft, and slippery terrains, our approach shows superior state estimation performance to the existing model-based state estimator. We summarize our contributions as follows.

% Under challenging scenarios including A, B, and C as well as flat, our approach shows reliability superior to a model-based estimator with strict contact assumption.

\begin{itemize}
\item We propose a novel proprioceptive state estimator that integrates a Neural Measurement Network (NMN) with IEKF.
\item We analyze the efficacy of existing learning techniques in reducing the sim-to-real gap of NMN learned only through simulation data.
\item We empirically demonstrate that NMN improves state estimation performance on flat, rough, soft, and slippery terrain.
\end{itemize}

\section{method}

In this section, we introduce a state estimation framework that integrates the Invariant Extended Kalman Filter (IEKF) with a Neural Measurement Network (NMN). This framework aims to make use of model-based and data-driven state estimators. The Schematic of the proposed framework is depicted in \figref{framework}. During the learning process, we sample trajectories through the locomotion policy in the simulation and train the NMN. In the inference process, we use the NMN as a measurement model in IEKF. Hereinafter, $\tilde{(\cdot)}$ represents the sensor value, ${(\cdot)}$ represents the true value, and $\hat{(\cdot)}$ represents the estimated value.

\subsection{Learning Process in Simulation}
\subsubsection{Architectures of Locomotion Policy and NMN}
We first train the locomotion policy of GRU~\cite{cho2014learning}-MLP architecture to control the quadruped robot. Thereafter, we freeze the policy network parameters and employ this policy throughout the training and inference processes.
\begin{figure}[tp]
\vspace{0.2cm}
\centering
\includegraphics[scale=0.72]{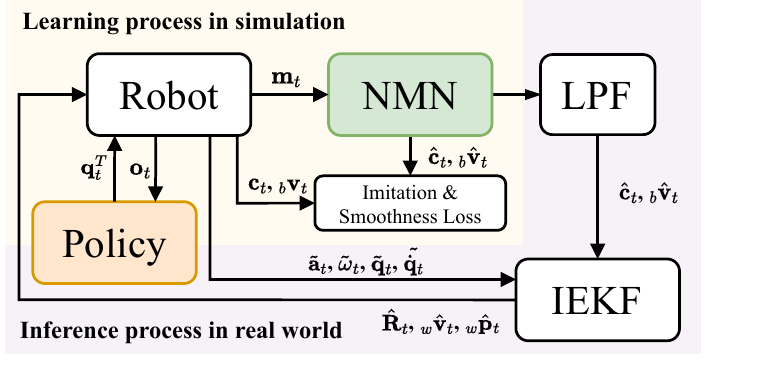}
% \includegraphics[width=\columnwidth]{figures/framework_comp.pdf}
% \vspace{-0.8cm}
\caption{\textbf{Schematic of the proposed framework:} To devise a learning-based state estimator, we first develop the policy for trajectory sampling. The NMN takes as input the proprioceptive sensor value and previous joint target and outputs contact probability and body linear velocity. These measurements are passed to the LPF and used to update the IEKF.}
\label{fig:framework}
% \vspace{-0.4cm}
\end{figure}
The observation of the locomotion policy is $\mathbf{o}_t \triangleq [ \tilde{\mathbf{g}}_t$, $\tilde{\mathbf{\omega}}_t$, $\tilde{\mathbf{q}}_t$, $\tilde{\dot{\mathbf{q}}}_t$, $\mathbf{q}^{des}_{t-1}$, $\mathbf{u}_t ] ^T$, where each component represents the roll-pitch vector of the body, body angular velocity, joint angles, joint velocities, previous positional joint targets, and velocity command at a given time $t$, respectively. Specifically, a vector $[ \tilde{\mathbf{g}}_t$, $\tilde{\mathbf{\omega}}_t$, $\tilde{\mathbf{q}}_t$, $\tilde{\dot{\mathbf{q}}}_t$, $\mathbf{q}^{des}_{t-1} ] ^T$ is first passed to the GRU and then concatenated with the hidden state of GRU and $\mathbf{u}_t$. The following MLP takes this stacked vector to produce new positional joint targets. \deleted{The roll-pitch vector is obtained directly from the IMU attached to the body.} \deleted{The policy is trained in the randomized terrain environment, as described by Lee et al.{~\cite{lee2020learning}}, utilizing the PPO{~\cite{schulman2017proximal}}. Training details such as action space, reward functions, and control frequency are nearly identical to the work of Choi et al.{~\cite{choi2023learning}}.} \added{To assess the standalone performance of the estimator, we intentionally separated it from the controller so that the policy observations can remain independent of states estimated.} \added{Therefore, the roll-pitch vector is obtained directly from the IMU attached to the body.} 

The NMN consists of the GRU-MLP architecture and measures $\hat{\mathbf{c}}_{t}$ and $_b\hat{\mathbf{v}}_{t}$, which are the contact probabilities of feet and the body linear velocity. Inputs to the NMN are $\mathbf{m}_t \triangleq [ \tilde{\mathbf{a}}_t$, $\tilde{\mathbf{\omega}}_t$, $\tilde{\mathbf{q}}_t$, $\tilde{\dot{\mathbf{q}}_t}$, $\mathbf{q}^{des}_{t-1} ] ^T$, where $\tilde{\mathbf{a}}_t$ represents acceleration. The hidden dimension of GRU was 128, and the MLP network has a $[256 \times 128]$ structure for both locomotion policy and the NMN.

\subsubsection{Simulation Details}
We train the control policy and the NMN using the Raisim~\cite{hwangbo2018per} simulator. The policy utilizing the PPO{~\cite{schulman2017proximal}} is trained in the randomized terrain environment, as described by Lee et al.{~\cite{lee2020learning}}. The terrain consists of flat, hills, steps, and stairs. Specifically, the amplitude of the hills follows $U(0.2, 1.4)$, the height of the steps follows $U(0.02, 0.18)~cm$, the height of the stairs follows $U(0.02, 0.18)~cm$, and the friction coefficient follows $U(0.4, 1.2)$. Additionally, to simulate unexpected slippage, we set the friction coefficient of the contact instance to the value sampled from $U(0.3, 0.4)$ with a 1\% probability. Training details such as action space, reward functions, and control frequency are nearly identical to the work of Choi et al.{~\cite{choi2023learning}}. 

The simulator and estimator operate at 4kHz and 500Hz, respectively. We generate trajectories through the pre-trained control policy and gather labels for the contact state and body linear velocity. For every iteration of the NMN training, all 400 environments are randomly initialized and generate a trajectory of 400 control steps. We inject zero mean Gaussian noise into $\mathbf{o}_t$ and $\mathbf{m}_t$ to simulate sensor noise. The noise ranges correspond to the specifications of the joint encoder and IMU. The simulation settings for the control policy training and the data collection are the same. \added{As with our control policy, reinforcement learning-based controllers often bridge the sim-to-real gap by introducing an extensive diversity to the domain properties, including robot kinematics or inertial parameters{~\cite{choi2023learning, ji2022concurrent}}. However, this strategy would be disadvantageous to the data-driven estimation scheme. We discover that applying domain randomization to our training can lower the velocity prediction accuracy once we have a sufficiently accurate robot model. We validate this result through an ablation study, which is detailed in the experimental results section.}

 \deleted{Reinforcement learning-based controllers often bridge the sim-to-real gap by introducing an extensive diversity to the domain properties, including robot kinematics or inertial parameters. However, this strategy would be ineffective or disadvantageous to the data-driven estimation scheme. We discover that introducing such randomization can lower the velocity prediction accuracy once we have a sufficiently accurate robot model. We validate this observation through an ablation study, which is detailed in the experimental results section.}

\subsubsection{Loss Functions}
Our loss function consists of two loss terms -- supervised learning loss $L_{sp}$ and regularization loss $L_{sm}$ for smoothness along the time axis, defined as follows:
\begin{equation} \label{loss function}
    \begin{split}
    &L_{sp} = L_{BCE}(\mathbf{c}_t, \hat{\mathbf{c}}_t) + L_{L1}(_b\mathbf{v}_t, _b\hat{\mathbf{v}}_t) \\
    &L_{sm} = \frac{1}{N} (||_b\hat{\mathbf{v}}_t - _b\hat{\mathbf{v}}_{t-1}||^2 + \frac{1}{2} ||_b\hat{\mathbf{v}}_{t} - 2_b\hat{\mathbf{v}}_{t-1} + _b\hat{\mathbf{v}}_{t-2} ||^2 )\\
    &L_{total} = L_{sp} + \lambda L_{sm}
    \end{split}
\end{equation}
The loss for supervised learning includes the binary cross-entropy loss for the contact probability and the mean absolute error loss for the body linear velocity. Inspired by Mysore et al.~\cite{mysore2021regularizing} and their CAPS proposal, we introduce the smoothness loss. This loss minimizes the first and second-order changes in the network's output to ensure a smoother result and narrow the sim-to-real gap. We apply the smoothness loss only to $_b\hat{\mathbf{v}}_{t}$ because the contact states are inherently discrete. In this context, $\lambda$ is a balancing factor between two loss terms, and we have set this to 50 throughout training. We further investigate the efficacy of the regularization in the experimental results section.

We use the Adam optimizer~\cite{kingma2014adam} for the training, with a learning rate of $5e-4$ and 32 epochs during the first 200 iterations. If the training continues further, the NMN would overfit the simulator and increase the sim-to-real gap. To avoid this problem, we use the early-stopping and the advantage of this technique is discussed in more detail in the experimental results section. 
% The training takes about 20 minutes on a standard desktop computer.

\subsection{Invariant Extended Kalman Filter}
\subsubsection{State Representation}
Our framework is built upon the IEKF for a legged robot, which is a world-centric right-invariant EKF proposed by Hartley et al.~\cite{hartley2020contact}. For ease of explanation, in this section, we describe the state space and system dynamics with the IMU biases omitted, but we implemented them in practice. The IMU frame coincides with the robot body frame. State variable $\mathbf{X}_t \in \mathrm{SE}_{N+2}(3)$ forms a matrix Lie group consisting of body state and $N$ contact points, which is defined as follows:
\begin{equation} \label{state}
    \mathbf{X_t} \triangleq 
    \begin{bmatrix}
        \mathbf{R_t} & \mathbf{v_t} & \mathbf{p_t} & \mathbf{d_t} \\
        \mathbf{0}_{1,3} & 1 & 0 & 0 \\
        \mathbf{0}_{1,3} & 0 & 1 & 0 \\
        \mathbf{0}_{1,3} & 0 & 0 & 1 \\
    \end{bmatrix}
\end{equation}
where $\mathbf{R}_t \in \mathrm{SO}(3), \mathbf{v}_t$, $\mathbf{p}_t$, and $\mathbf{d}_t$ represent the rotation matrix, linear velocity, position of the body, position of the contact point expressed in world frame, respectively. In the state matrix, $X_t$ can include multiple contact points. However, the system dynamics and measurement models for each contact point are independent. For better readability, we represent the state space for a single contact case.

\subsubsection{System Dynamics}
The IMU measurements are modeled as the sum of the true values and Gaussian noise, as follows:
\begin{equation} \label{IMU measurement}
    \begin{split}
    \mathbf{\tilde{\omega}}_t=\mathbf{\omega}_t+\mathbf{w}_t^g,~~\mathbf{w}_t^g \sim \mathcal{N}(\mathbf{0}_{3,1}, \Sigma^\mathbf{g}) \\
    \mathbf{\tilde{a}}_t=\mathbf{a}_t+\mathbf{w}_t^a,~~\mathbf{w}_t^a \sim \mathcal{N}(\mathbf{0}_{3,1}, \Sigma^\mathbf{a})
    \end{split}
\end{equation}
where $\mathcal{N}$ represents Gaussian noise. The position of the contact point remains stationary as long as the slip does not happen. We model the uncertainties on the contact position by adding Gaussian noise to the velocity of the contact point.
\begin{equation} \label{contact dynamics}
    _W\mathbf{\tilde{v}}_C=\mathbf{0}_{3,1}=_C\mathbf{v}_C+\mathbf{w}_t^v,~~\mathbf{w}_t^v \sim \mathcal{N}(\mathbf{0}_{3,1}, \Sigma^\mathbf{v})
\end{equation}
The system dynamics given the IMU and contact measurements are as follows:
\begin{equation} \label{system dynamics}
    \begin{split}
        &\frac{\mathrm{d} }{\mathrm{d} t}\mathbf{R}_t = \mathbf{R}_t (\tilde{\mathbf{\omega}}_t - \mathbf{w}_t^g)_\times \\
        &\frac{\mathrm{d} }{\mathrm{d} t}\mathbf{v}_t = \mathbf{R}_t (\tilde{\mathbf{a}}_t - \mathbf{w}_t^a) + \mathbf{g} \\
        &\frac{\mathrm{d} }{\mathrm{d} t}\mathbf{p}_t = \mathbf{v}_t \\
        &\frac{\mathrm{d} }{\mathrm{d} t}\mathbf{d}_t = \mathbf{R}_t \mathbf{h}_R(\tilde{\mathbf{q}}_t) (-\mathbf{w}_t^v)
    \end{split}
\end{equation}
where $(\cdot)_\times$ denotes a $3\times3$ skew-symmetric matrix and $\mathbf{h}_R(\tilde{\mathbf{q}}_t)$ is the orientation of the contact frame with respect to the body frame. This can be calculated from encoder measurements, $\tilde{\mathbf{q}}_t$, and forward kinematics.

\subsubsection{Right-Invariant Leg Kinematics Measurement Model}
To mimic the real sensor data, we further modeled the joint encoder measurements $\tilde{\mathbf{q}}_t$ with Gaussian noise as follows: 
\begin{equation} \label{encoder measurement}
    \tilde{\mathbf{q}}_t = \mathbf{q}_t+\mathbf{w}_t^{q},~~\mathbf{w}_t^q \sim \mathcal{N}(\mathbf{0}_{12,1}, \Sigma^\mathbf{q})
\end{equation}
The contact point under the non-slip conditions can be calculated through forward kinematics and expressed using state variables as follows:
\begin{equation} \label{forward kinematics position measurement}
\begin{split}
    &\mathbf{h}_p(\tilde{\mathbf{q}}_t) = \mathbf{R}_t^T(\mathbf{d}_t - \mathbf{p}_t) + \mathbf{J}_p(\tilde{\mathbf{q}}_t)\mathbf{w}_t^{q}
\end{split}
\end{equation}
where $\mathbf{h}_p(\tilde{\mathbf{q}}_t)$ is the forward kinematics of contact foot, and $\mathbf{J}_p(\tilde{\mathbf{q}}_t)$ is the analytical Jacobian of the forward kinematics. As Barrau et al.~\cite{barrau2016invariant} explained, the leg kinematics measurement model~{\eqref{forward kinematics position measurement}} follows right-invariant observation form, $\mathbf{Y}_t = \mathbf{X}_t^{-1}\mathbf{b} + \mathbf{V}_t$, where $\mathbf{Y}_t^T = \begin{bmatrix}\mathbf{h}_p^T(\tilde{\mathbf{q}}_t) & 0 & 1 & -1\end{bmatrix}$, $\mathbf{b}^T = \begin{bmatrix}\mathbf{0}_{1,3} & 0 & 1 & -1\end{bmatrix}$, $\mathbf{V}_t^T = \begin{bmatrix}(\mathbf{J}_p(\tilde{\mathbf{q}}_t)\mathbf{w}_t^{q})^T & 0 & 0 & 0\end{bmatrix}$.

\subsubsection{Right-Invariant Neural Measurement Model}
As shown in \figref{framework}, we construct the Neural Measurement Network (NMN) to measure contact probability $\hat{\mathbf{c}}_t$, and body linear velocity $_b\hat{\mathbf{v}}_t$. We utilize the learned-contact probability to determine the use of the leg kinematics measurement model. In addition, we formulate the learned-body-linear-velocity as follows:
\begin{equation} \label{body lin vel measurement}
    _b\hat{\mathbf{v}}_t = \mathbf{R}_t^T \mathbf{v}_t + \mathbf{w}_t^{NN},~~\mathbf{w}_t^{NN} \sim \mathcal{N}(\mathbf{0}_{3,1}, \Sigma^{NN})
\end{equation}
where $\mathbf{w}_t^{NN}$ is Gaussian noise that represents the uncertainty of neural measurement. Our learned-body-linear-velocity formulation~{\eqref{body lin vel measurement}} also satisfies the right-invariant observation form, where $\mathbf{Y}_t^T = \begin{bmatrix}_b\hat{\mathbf{v}}_t & -1 & 0 & 0\end{bmatrix}$, $\mathbf{b}^T = \begin{bmatrix}\mathbf{0}_{1,3} & -1 & 0 & 0\end{bmatrix}$, $\mathbf{V}_t^T = \begin{bmatrix}(\mathbf{w}_t^{NN})^T & 0 & 0 & 0\end{bmatrix}$.

\subsubsection{Kalman Update}
By utilizing the log-linear property of the right-invariant error and first-order approximation, full state and covariance update equations of the IEKF can be formulated as follows:
\begin{equation} \label{Kalman update}
    \begin{split}
    &\bar{\mathbf{X}}_t^+ = \mathrm{exp}(\mathbf{K}_t \Pi \bar{\mathbf{X}}_t \mathbf{Y}_t) \bar{\mathbf{X}}_t \\
    &\mathbf{P}_t^+ = (\mathbf{I} - \mathbf{K}_t \mathbf{H}_t)\mathbf{P}_t(\mathbf{I} - \mathbf{K}_t \mathbf{H}_t)^T + \mathbf{K}_t \bar{\mathbf{N}}_t \mathbf{K}_t^T
    \end{split}
\end{equation}
where $\bar{\mathbf{X}}_t$ is the estimated state from system dynamics, $\mathbf{P}_t$ is the covariance matrix, and $\Pi \triangleq \begin{bmatrix} \mathbf{I} & \mathbf{0}_{3,3}\end{bmatrix}$ is the auxiliary selection matrix. Kalman gain $\mathbf{K}_t$ is computed as the following:
\begin{equation} \label{Kalman gain}
    \mathbf{S}_t = \mathbf{H}_t \mathbf{P}_t \mathbf{H}_t^T + \bar{\mathbf{N}}_t ,~~ \mathbf{K}_t =  \mathbf{P}_t  \mathbf{H}_t^T  \mathbf{S}_t^{-1}
\end{equation}
We set the observation model as $\mathbf{H}_t = \begin{bmatrix} ^q\mathbf{H}_t^T & ^v\mathbf{H}_t^T \end{bmatrix}^T$ and $\mathbf{N}_t = \begin{bmatrix} ^q\mathbf{N}_t & \mathbf{0}_{3,3}; & \mathbf{0}_{3,3} & ^v\mathbf{N}_t \end{bmatrix}$, where $^q(\cdot)$ is for the leg kinematics model, and $^v(\cdot)$ represents the learned-body-linear-velocity model. The observation model for the leg kinematics measurement can be drawn from the linear update equation as follows:
\begin{equation} \label{H and N for contact measurement}
    \begin{split}
    &^q\mathbf{H}_t = \begin{bmatrix}\mathbf{0}_{3,3} & \mathbf{0}_{3,3} & -\mathbf{I} & \mathbf{I}\end{bmatrix} \\
    &^q\bar{\mathbf{N}}_t = \bar{\mathbf{R}}_t \mathbf{J}_p(\tilde{\mathbf{q}}_t) \mathrm{Cov}(\mathbf{w}_t^{q}) \mathbf{J}_p^T(\tilde{\mathbf{q}}_t) \bar{\mathbf{R}}_t^T
    \end{split}
\end{equation}
The observation model for the learned-body-linear-velocity measurement is as follows:
\begin{equation} \label{H and N for body lin vel measurement}
    \begin{split}
    &^v\mathbf{H}_t = \begin{bmatrix}\mathbf{0}_{3,3} & \mathbf{I} & \mathbf{0}_{3,3} & \mathbf{0}_{3,3}\end{bmatrix} \\
    &^v\bar{\mathbf{N}}_t = \bar{\mathbf{R}}_t \mathrm{Cov}(\mathbf{w}_t^{NN}) \bar{\mathbf{R}}_t^T
    \end{split}
\end{equation}

We refer the reader to Hartley et al.~\cite{hartley2020contact} for more details on the material described above and additional IMU bias augmentation explanation.

\subsection{Inference Process in Real World}
During the inference process, we set the covariance as follows:
\begin{equation} \label{covariance}
    \begin{split}
    &\mathrm{Cov}(\mathbf{w}_t^g, \mathbf{w}_t^a, \mathbf{w}_t^v, \mathbf{w}_t^{q}, \mathbf{w}_t^{NN}, \mathbf{w}_t^{b_g}, \mathbf{w}_t^{b_a}) \\
    &= (10^{-5}, 10^{-1}, 10^{-4},10^{-6}, 10^{-5.5}, 10^{-10}, 10^{-10})\mathbf{I}_{3,3} \\
    &\mathrm{Cov}(\mathbf{R}_0, \mathbf{v}_0, \mathbf{p}_0, \mathbf{b}_{g_0}, \mathbf{b}_{a_0}) \\
    &= (10^{-8}, 10^{-8}, 10^{-8},10^{-10}, 10^{-10})\mathbf{I}_{3,3}
    \end{split}
\end{equation}
where $\mathbf{w}_{b_g}, \mathbf{w}_{b_a}$ represent bias uncertainties. The true uncertainty regarding the contact position should be $\mathbf{h}_R(\tilde{\mathbf{q}}_t) (-\mathbf{w}_t^v)$, which is indicated in \eqref{system dynamics}. However, due to the unknown orientation of the contact frame, we approximate this to a constant value. We additionally utilize a first-order low-pass filter~(LPF) in the inference process with cutoff frequencies of 40Hz for $\hat{\mathbf{c}}_t$ and 10Hz for $_b\hat{\mathbf{v}}_{t}$. We threshold $\hat{\mathbf{c}}_t$ to 0.5 to determine the contact state and use it in IEKF. We have found that although the robot remains stationary, the learned-body-linear-velocity measurement of the NMN has a bias with a precision on the order of $10^{-3}$. To prevent position drift caused by this bias, we apply \eqref{body lin vel measurement} as the measurement model only when $\hat{\mathbf{v}}_t$ exceeds 0.1.
\begin{figure*}[htbp]
\vspace{0.2cm}
\centering
\includegraphics[scale=0.268]{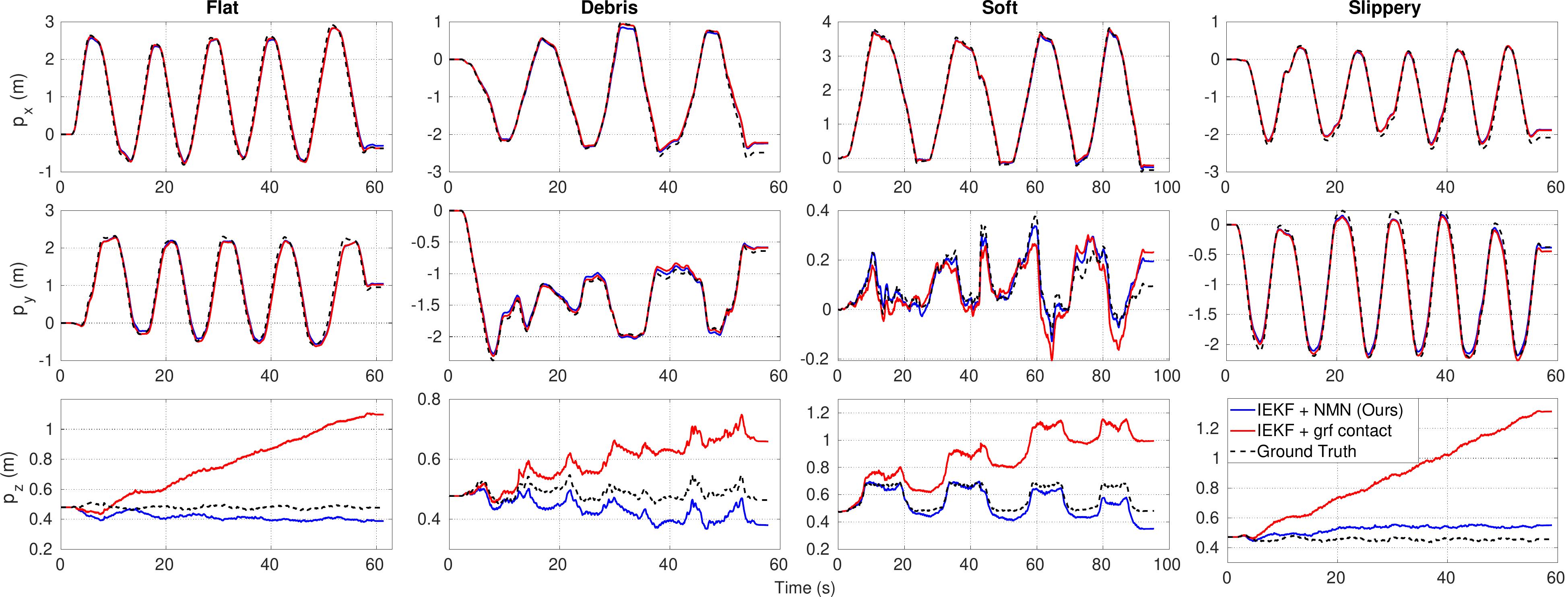}
\caption{\textbf{Position estimation comparison:} Proposed NMN vs. model-based GRF contact on four different terrains. Our NMN reduces the z-direction position drift compared to the model-based approach. Either method employs the slip rejection method.}
\label{fig:est result}
% \vspace{-0.3cm}
\end{figure*}

\section{Experimental Results}
We design experiments to investigate the following research questions: 
(1) Can our framework improve the state estimation performance using only simulation data? (2) How can we reduce the sim-to-real gap in our supervised approach?

We gather sensor measurements using Raibo and process these measurements offline through our framework. The IMU (Microstrain-3DM) delivers data at 166 Hz, and encoder measurements are recorded at 500 Hz. To calculate the estimation error, we collect ground truth data from a motion capture system (12 Vicon Vero V2.2 cameras) at 250Hz and then interpolate the motion capture data to 500Hz for compatibility with our estimation results. We align the initial state with the ground truth and set the IMU bias to zero.

\begin{figure}[t]
\centering
\includegraphics[scale=0.52]{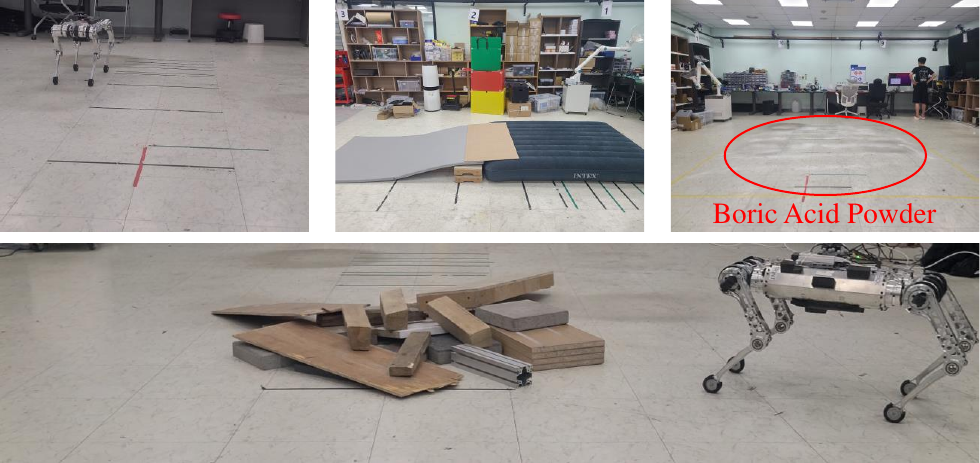}
\caption{\textbf{Experiment Terrain Overview:} Upper left - \textit{Flat}, Upper middle - \textit{Soft}, Upper right - \textit{Slippery}, Lower - \textit{Debris}.}
\label{fig:experiment terrain figure}
% \vspace{-0.4cm}
\end{figure}

\subsection{Experimental Details}
\subsubsection{Baselines}
We select four baseline estimators as follows:
\begin{itemize}
    % \item \textbf{IEKF with ground reaction force (GRF) contact:} 
    \item \textbf{ground reaction force (GRF) contact:} 
    % It calculates the GRF through the dynamics model and determines the contact state by comparing its magnitude with the predefined threshold, which is 40N. The estimated contact state is then passed to the IEKF. We smooth the GRF with an LPF with a cutoff frequency of 10 Hz.
    % This state estimator utilizes the leg kinematics measurement model with IEKF. In this estimator, GRF is calculated from robot dynamics and filtered through a 10Hz LPF. The contact state is determined by thresholding the filtered GRF at 40N and used in IEKF for the leg kinematics measurement.
    The first baseline estimator is based on IEKF and utilizes the contact state calculated through the GRF for the leg kinematics measurement. In this estimator, 
    % we calculate the GRF from robot dynamics and filter it through a 10Hz LPF. 
    we calculate the GRF from robot dynamics and filter it through an LPF with a cut-off frequency of 10Hz.
    The contact state is determined by thresholding the filtered GRF at 40N.
    
    \item \textbf{IEKF with NMN $\hat{\mathbf{c}}_t$:} 
    % It activates or deactivates the leg kinematics measurement model according to the $\hat{\mathbf{c}}_t$ predicted by NMN, but it does not utilize the predicted $_b\hat{\mathbf{v}}_t$ for the Kalman update, i.e., $\mathbf{H}_t = ^q\mathbf{H}_t$ and $\bar{\mathbf{N}}_t = ^q\bar{\mathbf{N}}_t$.
    % This state estimator is based on the contact-aid IEKF. In this estimator, the contact state is determined from the $\hat{\mathbf{c}}_t$ predicted by NMN and used in IEKF as the leg kinematics measurement. This estimator does not utilize the predicted $_b\hat{\mathbf{v}}_t$ for the Kalman update, i.e., $\mathbf{H}_t = ^q\mathbf{H}_t$ and $\bar{\mathbf{N}}_t = ^q\bar{\mathbf{N}}_t$.
    The second baseline estimator is based on IEKF and utilizes the $\hat{\mathbf{c}}_t$ predicted by NMN for the leg kinematics measurement. This estimator does not utilize the predicted $_b\hat{\mathbf{v}}_t$ for the Kalman update, i.e., $\mathbf{H}_t = ^q\mathbf{H}_t$ and $\bar{\mathbf{N}}_t = ^q\bar{\mathbf{N}}_t$.
    
    \item \textbf{IEKF with NMN $_b\hat{\mathbf{v}}_t$:} 
    % It utilizes the body linear velocity measurement model using the $_b\hat{\mathbf{v}}_t$ predicted by NMN, but it does not employ the predicted $\hat{\mathbf{c}}_t$, i.e., $\mathbf{H}_t = ^v\mathbf{H}_t$ and $\bar{\mathbf{N}}_t = ^v\bar{\mathbf{N}}_t$.
    The third baseline estimator is based on IEKF and utilizes the $_b\hat{\mathbf{v}}_t$ predicted by NMN for the body linear velocity measurement. This estimator does not utilize the predicted $\hat{\mathbf{c}}_t$ for the Kalman update, i.e., $\mathbf{H}_t = ^v\mathbf{H}_t$ and $\bar{\mathbf{N}}_t = ^v\bar{\mathbf{N}}_t$.

    % This state estimator utilizes the $_b\hat{\mathbf{v}}_t$ predicted by NMN with IEKF.
    
    \item \textbf{NN only:} The fourth baseline estimator is a GRU-MLP network trained using end-to-end regression method. We train the GRU-MLP network to directly estimate relative position $\mathbf{R}_{t-1}^T(\mathbf{p}_t - \mathbf{p}_{t-1})$, orientation $Log(\mathbf{R}_{t-1 }^T \mathbf{R}_t)$, and linear velocity $_b\hat{\mathbf{v}}_t$. The estimation result is calculated by integrating the network output. Jin et al.~\cite{jin2023learning} described that learning the relative pose can be inconsistent when only the IMU measurements are available. However, we assume that access to leg kinematics can circumvent such problems by making $_b\hat{\mathbf{v}}_t$ and $\mathbf{g}_t$ to be observable.
\end{itemize}

\subsubsection{Slip rejection}
% Our framework is orthogonal to the slip rejection method proposed by Kim et al.~\cite{kim2021legged}, and we further study how it performs within our framework. 
The slip rejection method proposed by Kim et al.~\cite{kim2021legged} is suitable for use with any state estimators using leg kinematics measurement. 
Since our estimator utilizes leg kinematics measurement, we can apply this method to our estimator and study how it performs.
% We apply this method to our estimator and study how it performs, given that our estimator utilizes leg kinematic measurements.

% The rejection method detects foot slippage when the velocity of the contact foot is larger than the threshold which we set to 0.4m/s. For the detected slip, it augments the covariance of the contact foot velocity $\mathrm{Cov}(\mathbf{w}_t^v)$ by a factor of ten so that the Kalman update can reflect the uncertainties in the leg kinematics model.
In the rejection method, we detect foot slip when the velocity of the contact foot is larger than the threshold, which we set to 0.4m/s. When the slip is detected, we increase the covariance of the contact foot velocity $\mathrm{Cov}(\mathbf{w}_t^v)$ by a factor of ten so that the uncertainties of the leg kinematics model are reflected in the Kalman update.

\begin{table}[t]
\vspace{0.2cm}
\caption{ATE and 10-second RE on State Estimation Results Across Various Environment Experiments}
\label{tab:estimation result}
\renewcommand{\arraystretch}{1.5}
\setlength{\tabcolsep}{5pt}
\resizebox{\columnwidth}{!}
{
\begin{tabular}{|c|c|c|c|c|c|c|c|c|}
\hline
\textbf{Terrain}           & \textbf{\begin{tabular}[c]{@{}c@{}}Slip\\ rejection\end{tabular}} 
                           & \textbf{Method}   
                           & \textbf{\begin{tabular}[c]{@{}c@{}}ATE\\ (pos)\end{tabular}} 
                           & \textbf{\begin{tabular}[c]{@{}c@{}}ATE\\ (vel)\end{tabular}} 
                           & \textbf{\begin{tabular}[c]{@{}c@{}}ATE\\ (ori)\end{tabular}} 
                           & \textbf{\begin{tabular}[c]{@{}c@{}}RE\\ (pos)\end{tabular}} 
                           & \textbf{\begin{tabular}[c]{@{}c@{}}RE\\ (vel)\end{tabular}} 
                           & \textbf{\begin{tabular}[c]{@{}c@{}}RE\\ (ori)\end{tabular}} \\ \hline

\multirow{8}{*}{\textit{Flat}}     & \multirow{5}{*}{Off} & \textbf{Proposed}    & \textbf{0.2110} & 0.0759 & 0.0824 & 0.2350 & \textbf{0.0972} & 0.0396 \\ \cline{3-9} 
                           &                      & w/ $_b\hat{\mathbf{v}}_t$ & 0.2306 & 0.0781 & 0.0824 & 0.2351 & 0.0983 &	0.0399 \\ \cline{3-9} 
                           &                      & w/ $\hat{\mathbf{c}}_t$   & 0.3585 & 0.0791 & 0.0824 & 0.2418 & 0.1035 &	0.0397 \\ \cline{3-9} 
                           &                      & GRF contact    & 0.4346 & \textbf{0.0747} & 0.0817 & 0.2399 & 0.0995 &	0.0397 \\ \cline{3-9} 
                           &                      & NN only             & 0.8142 & 0.1212 & 0.4100 & 0.8674 & 0.1727 &	0.1121 \\ \cline{2-9} 
                           
                           & \multirow{3}{*}{On}  & \textbf{Proposed}   & 0.2142 & 0.0760 & 0.0814 & \textbf{0.2332} & 0.0972 & \textbf{0.0396} \\ \cline{3-9} 
                           &                      & w/ $\hat{\mathbf{c}}_t$   & 0.3169 & 0.0789 & 0.0817 & 0.2390 & 0.1040 & 0.0398 \\ \cline{3-9} 
                           &                      & GRF contact    & 0.3459 & 0.0770 & \textbf{0.0809} & 0.2363 & 0.1031 & 0.0398 \\ \hline

\multirow{8}{*}{\textit{Debris}}   & \multirow{5}{*}{Off} & \textbf{Proposed}    & 0.1798 & 0.0877 & \textbf{0.0591} & 0.1166 & 0.1281 & 0.0625 \\ \cline{3-9} 
                           &                      & w/ $_b\hat{\mathbf{v}}_t$ & 0.1626 & 0.0917 & 0.0610 & 0.1173 & 0.1299 & 0.0618 \\ \cline{3-9} 
                           &                      & w/ $\hat{\mathbf{c}}_t$   & 0.2630 & 0.1030 & 0.0597 & 0.1321 & 0.1563 & 0.0618 \\ \cline{3-9} 
                           &                      & GRF contact    & 0.3074 & 0.0884 & 0.0604 & 0.1087 & 0.1325 & \textbf{0.0615} \\ \cline{3-9} 
                           &                      & NN only             & 0.3983 & 0.1254 & 0.0784 & 0.1845 & 0.1932 & 0.1000 \\ \cline{2-9} 
                           
                           & \multirow{3}{*}{On}  & \textbf{Proposed}   & 0.1738 & 0.0875 & 0.0599 & 0.1165 & 0.1280 & 0.0619 \\ \cline{3-9} 
                           &                      & w/ $\hat{\mathbf{c}}_t$   & \textbf{0.1290} & 0.0882 & 0.0601 & 0.1020 & 0.1292 & 0.0617 \\ \cline{3-9} 
                           &                      & GRF contact    & 0.1478 & \textbf{0.0848} & 0.0608 & \textbf{0.0933} & \textbf{0.1261} & 0.0617 \\ \hline

\multirow{8}{*}{\textit{Soft}}     & \multirow{5}{*}{Off} & \textbf{Proposed}    & 0.2450 & \textbf{0.0799} & 0.0296 & \textbf{0.0772} & \textbf{0.1180} & 0.0404 \\ \cline{3-9} 
                           &                      & w/ $_b\hat{\mathbf{v}}_t$ & 0.2313 & 0.0835 & 0.0452 & 0.1031 & 0.1200 & 0.0408 \\ \cline{3-9} 
                           &                      & w/ $\hat{\mathbf{c}}_t$   & 0.2251 & 0.0988 & \textbf{0.0282} & 0.1218 & 0.1451 & \textbf{0.0398} \\ \cline{3-9} 
                           &                      & GRF contact    & 0.7228 & 0.0858 & 0.0386 & 0.1282 & 0.1238 & 0.0402 \\ \cline{3-9} 
                           &                      & NN only             & 0.5087 & 0.1035 & 0.2266 & 0.3209 & 0.1551 & 0.1269 \\ \cline{2-9} 
                           
                           & \multirow{3}{*}{On}  & \textbf{Proposed}   & \textbf{0.2059} & 0.0800 & 0.0362 & 0.0869 & 0.1182 & 0.0407 \\ \cline{3-9} 
                           &                      & w/ $\hat{\mathbf{c}}_t$   & 0.3093 & 0.0819 & 0.0361 & 0.1059 & 0.1180 & 0.0401 \\ \cline{3-9} 
                           &                      & GRF contact    & 0.3115 & 0.0820 & 0.0401 & 0.1004 & 0.1196 & 0.0403 \\ \hline

\multirow{8}{*}{\textit{Slippery}} & \multirow{5}{*}{Off} & \textbf{Proposed}    & 0.1792 & 0.0694 & \textbf{0.0467} & \textbf{0.1421} & 0.0774 & \textbf{0.0199} \\ \cline{3-9} 
                           &                      & w/ $_b\hat{\mathbf{v}}_t$ & 0.1751 & 0.0715 & 0.0675 & 0.1619 & 0.0783 & 0.0234 \\ \cline{3-9} 
                           &                      & w/ $\hat{\mathbf{c}}_t$   & 0.5916 & 0.1016 & 0.0511 & 0.2075 & 0.1246 & 0.0216 \\ \cline{3-9} 
                           &                      & GRF contact    & 0.6675 & 0.0931 & 0.0527 & 0.2114 & 0.1151 & 0.0215 \\ \cline{3-9} 
                           &                      & NN only             & 0.4223 & 0.1015 & 0.2971 & 0.4018 & 0.1448 & 0.0747 \\ \cline{2-9} 
                           
                           & \multirow{3}{*}{On}  & \textbf{Proposed}   & \textbf{0.1623} & \textbf{0.0693} & 0.0509 & 0.1432 & \textbf{0.0773} & 0.0208 \\ \cline{3-9} 
                           &                      & w/ $\hat{\mathbf{c}}_t$   & 0.3807 & 0.0792 & 0.0526 & 0.1558 & 0.0948 & 0.0216 \\ \cline{3-9} 
                           &                      & GRF contact    & 0.4412 & 0.0729 & 0.0528 & 0.1556 & 0.0876 & 0.0215 \\ \hline

\multirow{8}{*}{Total Average} & \multirow{5}{*}{Off} & \textbf{Proposed}     & 0.2038 & 0.0782 & \textbf{0.0545} & \textbf{0.1427} & \textbf{0.1052} & \textbf{0.0406} \\ \cline{3-9} 
                           &                      & w/ $_b\hat{\mathbf{v}}_t$ & 0.1999 & 0.0812 & 0.0640 & 0.1544 & 0.1066 & 0.0415 \\ \cline{3-9} 
                           &                      & w/ $\hat{\mathbf{c}}_t$   & 0.3596 & 0.0956 & 0.0554 & 0.1758 & 0.1324 & 0.0407 \\ \cline{3-9} 
                           &                      & GRF contact          & 0.5331 & 0.0855 & 0.0584 & 0.1721 & 0.1177 & 0.0407 \\ \cline{3-9} 
                           &                      & NN only                   & 0.5359 & 0.1129 & 0.2530 & 0.4437 & 0.1665 & 0.1034 \\ \cline{2-9} 
                           
                           & \multirow{3}{*}{On}  & \textbf{Proposed}         & \textbf{0.1891} & \textbf{0.0782} & 0.0571 & 0.1450 & 0.1052 & 0.0408 \\ \cline{3-9} 
                           &                      & w/ $\hat{\mathbf{c}}_t$   & 0.2840 & 0.0821 & 0.0576 & 0.1507 & 0.1115 & 0.0408 \\ \cline{3-9} 
                           &                      & GRF contact          & 0.3116 & 0.0792 & 0.0587 & 0.1464 & 0.1091 & 0.0408 \\ \hline
                           
\end{tabular}
}
% \vspace{-0.3cm}
\end{table}

\subsection{Hardware Experiments}
We test four different terrains to evaluate the proposed NMN and IEKF algorithms: \textit{flat, debris, soft}, and \textit{slippery}. \figref{experiment terrain figure} illustrates the experimental setup. 
\begin{itemize}
    \item \textbf{\textit{Flat}:} 
    The robot traverses a circular trajectory of 49 meters of flat ground for 62 seconds with a friction coefficient of over 0.6.
    \item \textbf{\textit{Debris}:} 
    We simulate the debris terrain with randomly stacked wooden logs and bricks, and the robot repeatedly navigates obstacles over 26 meters for 59 seconds.
    \item \textbf{\textit{Soft}:} 
    The terrain consists of thin mats and air mats, and the robot makes four round trips of 33 meters for 98 seconds.
    \item \textbf{\textit{Slippery}:} 
    We disperse the boric acid powder on a flat surface, lowering the coefficient of friction to less than 0.3. The robot traverses a circular trajectory of 45 meters for 59 seconds.
\end{itemize}
Robots are often faced with ambiguous contact situations, except for the \textit{flat}.

\begin{table}[t]
\vspace{0.2cm}
\caption{\textbf{Ablation Study:} Impact of Smoothness Loss and Domain Randomization on Various Environment Experiments}
\label{tab:ablation}
\renewcommand{\arraystretch}{1.5}
\setlength{\tabcolsep}{5pt}
\resizebox{\columnwidth}{!}
{
\begin{tabular}{|c|c|c|c|c|c|c|c|}
\hline
\textbf{Terrain}           & \textbf{Method}   
                           & \textbf{\begin{tabular}[c]{@{}c@{}}ATE\\ (pos)\end{tabular}} 
                           & \textbf{\begin{tabular}[c]{@{}c@{}}ATE\\ (vel)\end{tabular}} 
                           & \textbf{\begin{tabular}[c]{@{}c@{}}ATE\\ (ori)\end{tabular}} 
                           & \textbf{\begin{tabular}[c]{@{}c@{}}RE\\ (pos)\end{tabular}} 
                           & \textbf{\begin{tabular}[c]{@{}c@{}}RE\\ (vel)\end{tabular}} 
                           & \textbf{\begin{tabular}[c]{@{}c@{}}RE\\ (ori)\end{tabular}} \\ \hline

\multirow{3}{*}{\textit{Flat}}     & \textbf{Proposed} & \textbf{0.2110} & \textbf{0.0759} & 0.0824 & \textbf{0.2350} & \textbf{0.0972} & 0.0396 \\ \cline{2-8}
                          & w/o smoothness    & 0.4592 & 0.0891 & 0.0820 & 0.2455 & 0.1120 & 0.0393 \\ \cline{2-8}
                          & w/ randomization & 0.3556 & 0.0864 & \textbf{0.0786} & 0.2543 & 0.0992 & \textbf{0.0389} \\ \hline

\multirow{3}{*}{\textit{Debris}}   & \textbf{Proposed} & \textbf{0.1798} & \textbf{0.0877} & 0.0591 & \textbf{0.1166} & \textbf{0.1281} & 0.0625 \\ \cline{2-8}
                          & w/o smoothness    & 0.3650 & 0.0981 & \textbf{0.0589} & 0.1480 & 0.1418 & 0.0629 \\ \cline{2-8}
                          & w/ randomization & 0.3121 & 0.0934 & 0.0603 & 0.1579 & 0.1325 & \textbf{0.0623} \\ \hline

\multirow{3}{*}{\textit{Soft}}     & \textbf{Proposed} & 0.2450 & \textbf{0.0799} & 0.0296 & 0.0772 & \textbf{0.1180} & 0.0404\\ \cline{2-8}
                          & w/o smoothness    & \textbf{0.1338} & 0.0874 & 0.0305 & \textbf{0.0771} & 0.1285 & 0.0409 \\ \cline{2-8}
                          & w/ randomization & 0.4432 & 0.0825 & \textbf{0.0285} & 0.1165 & 0.1205 & \textbf{0.0398} \\ \hline

\multirow{3}{*}{\textit{Slippery}} & \textbf{Proposed} & 0.1792 & \textbf{0.0694} & \textbf{0.0467} & \textbf{0.1421} & \textbf{0.0774} & \textbf{0.0199} \\ \cline{2-8}
                          & w/o smoothness    & \textbf{0.1785} & 0.0845 & 0.0491 & 0.1644 & 0.1021 & 0.0201 \\ \cline{2-8}
                          & w/ randomization & 0.1913 & 0.0882 & 0.0498 & 0.1788 & 0.0823 & 0.0206 \\ \hline
                           
\end{tabular}
}
% \vspace{-0.4cm}
\end{table}

\subsubsection{Estimation results}
The estimated positions for each terrain are shown in \figref{est result}. Compared to the baseline, NMN shows a substantially small z-direction drift. The quantitative estimation results are presented in \tabref{estimation result}. In this table, we highlight the best-performing metrics in bold.
For methodologies employing Neural Networks (NN), we train the network from five different seeds and average the outcomes. The definitions for Absolute Trajectory Error (ATE) and Relative Error (RE) refer to~\cite{zhang2018tutorial}, where RE is computed with 10-second sub-trajectories. The error units are $\text{rad}, ~\text{m/s},~\text{m}$ for rotation, velocity, and position, respectively. 

\textbf{IEKF with NMN $\hat{\mathbf{c}}_t$} method shows a lower $\text{ATE}_{pos}$ and a higher $\text{ATE}_{vel}$ than \textbf{IEKF with GRF contact} method. \added{The difference between the two methods is trivial, indicating that the learned-contact is similar to that calculated through GRF.}
We suspect this \added{small} difference stems from the simulator's tendency to overestimate contact. The learned-contact is often more sensitive than the GRF thresholding scheme because, in the training environment, we can identify the contact even with a small interaction force and provide the simulated contact state as the ground-truth label.
\textbf{IEKF with NMN $_b\hat{\mathbf{v}}_t$} method has generally performed well among the estimator, especially for the terrains that cause unstable contact.
 This result suggests that state estimators for legged robots can improve estimation performance through the body linear velocity measurement model.

We find that our proposed method yields impressive outcomes for \textit{soft}, which is not covered in training scenarios. While the slip rejection method does bolster the accuracy of estimations based on leg kinematics measurements, our strategy outperforms it in many cases.
The \textbf{NN only} method shows high estimation errors across different terrains. Overall, we find that the NMN using only simulation data can improve the state estimation performance and exhibit adaptability to challenging terrains where unstable contacts occur, even including contact scenarios not previously encountered.

\subsubsection{Ablation study}

In this section, we delve into the effects of the smoothness loss, domain randomization, and early stopping, which are the techniques we introduced to mitigate the sim-to-real gap. 
% We first train a network without the smoothness loss and another network with domain randomization while the rest of the training setup remains unchanged.
Before conducting the ablation study, we would like to clarify key aspects of the proposed method. Our proposed method uses the smoothness loss function and does not employ domain randomization during the learning process.
We conduct an ablation study on smoothness loss function and domain randomization.

\begin{table}[t]
\vspace{0.2cm}
\caption{\textbf{Ablation Study:} Average Standard Deviation of Linear Velocity Error on Flat Terrain Experiment}
\label{tab:smoothness}
\renewcommand{\arraystretch}{1.2}
\setlength{\tabcolsep}{10pt}
\resizebox{\columnwidth}{!}
{
\begin{tabular}{|c|c|c|c|c|}
\hline
\textbf{Terrain}           & \textbf{Method}   
                           & \textbf{\begin{tabular}[c]{@{}c@{}}$v_x$\end{tabular}} 
                           & \textbf{\begin{tabular}[c]{@{}c@{}}$v_y$\end{tabular}} 
                           & \textbf{\begin{tabular}[c]{@{}c@{}}$v_z$\end{tabular}} \\ \hline

\multirow{2}{*}{\textit{Flat}}     & \textbf{w/} ${L_{sm}}$       & \textbf{0.0472} & \textbf{0.0386} & \textbf{0.0435} \\ \cline{2-5}
                          % & 1000 iteration               & 0.0482 & 0.0386 & 0.0424 \\ \cline{2-5}
                          & w/o $L_{sm}$               & 0.0540 & 0.0466 & 0.0499 \\ \hline
                           
\end{tabular}
}
\end{table}

% \hspace{1.5cm}
\begin{figure}[t]
\centering
\includegraphics[width=\columnwidth]{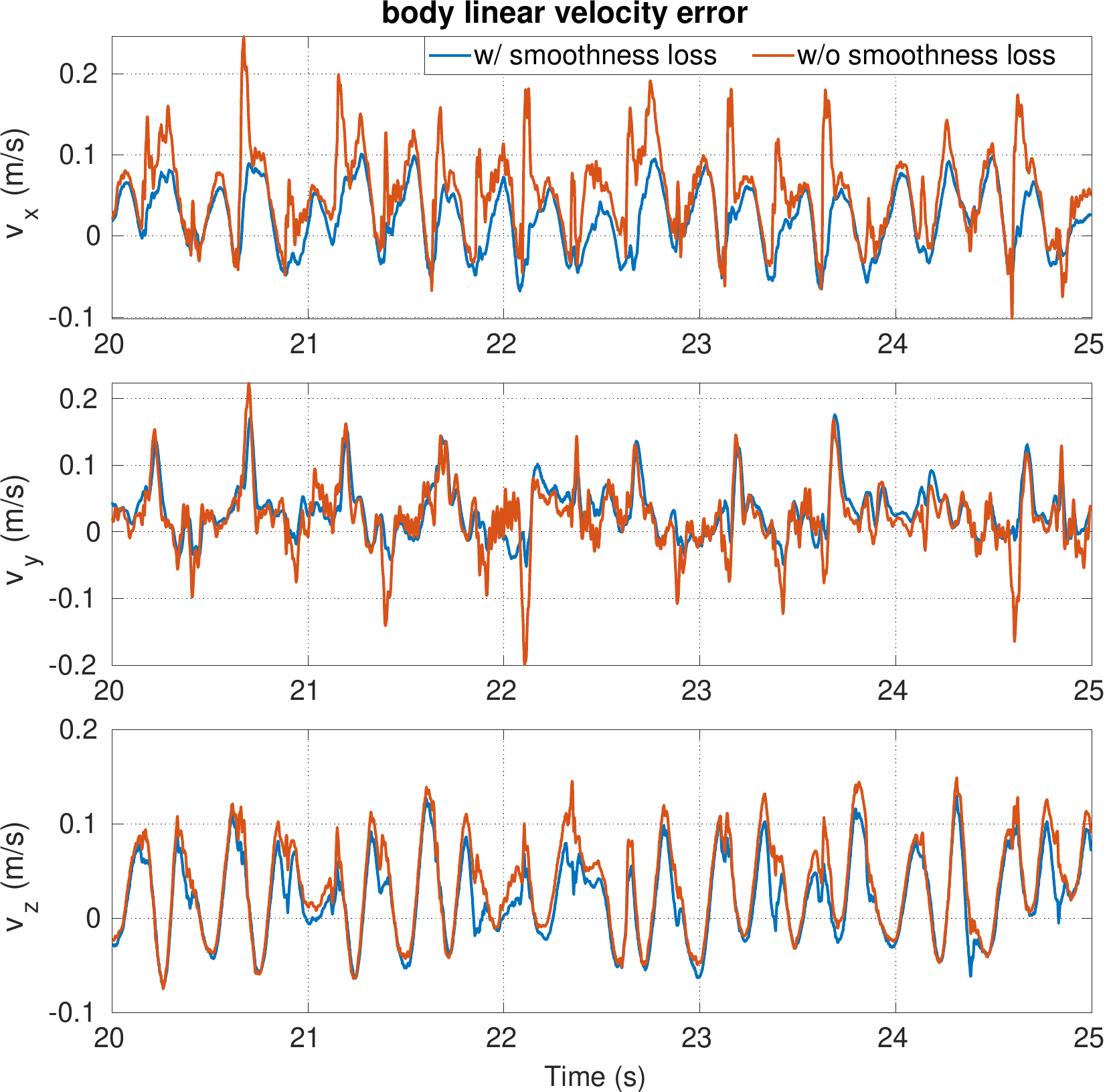}
\caption{We compare the body linear velocity error on the \textit{flat} experiment for each ablation study.}
\label{fig:body lin vel error}
% \vspace{-0.5cm}
\end{figure}

The estimation results of the ablation study are presented in \tabref{ablation}. Our proposed method outperforms the variant estimators for most cases. For the ablation study on smoothness loss, the contact state is set the same as the proposed method. In the case without smoothness loss, the estimation accuracy decreases drastically. We also find that the error's standard deviation increases in \tabref{smoothness}, and the network output appears to have more fluctuation and peaks in \figref{body lin vel error}. For the ablation study on domain randomization, applying domain randomization degenerates the estimation performance for most cases across different terrains. This implies that domain randomization might exacerbate the sim-to-real gap for our estimation framework, unlike the usual sim-to-real transfer in reinforcement learning-based controllers.

\begin{figure}[t]
\centering
\includegraphics[width=\columnwidth]{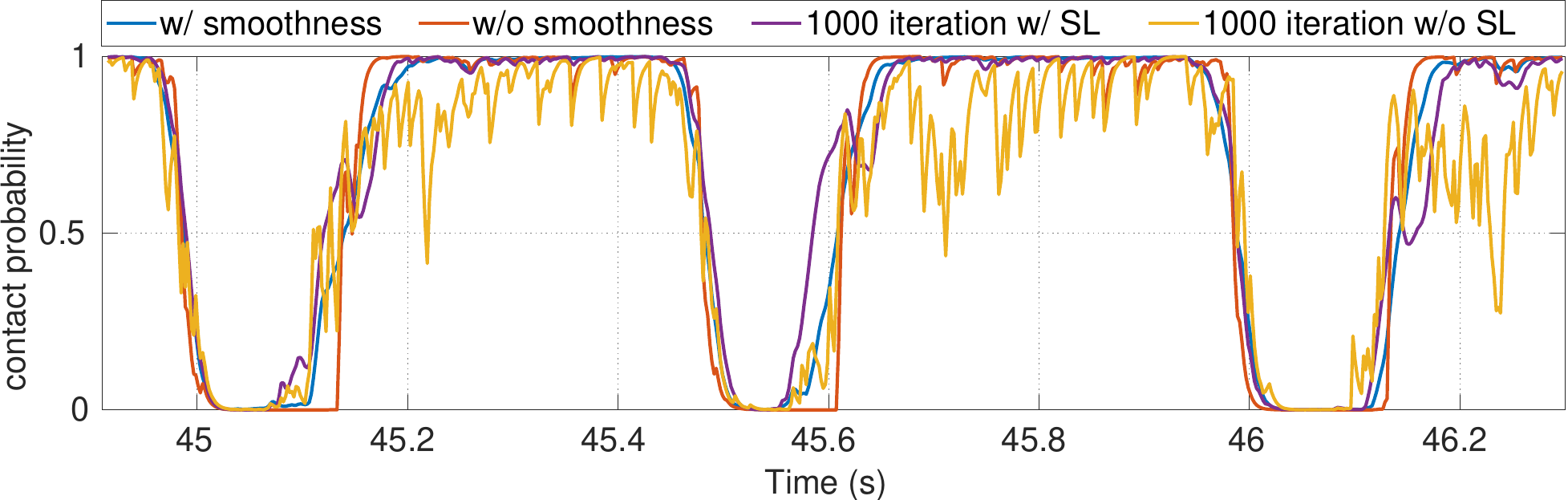}
\caption{We compare the learned-contact probability that processes it through the LPF on the \textit{flat} experiment for each ablation study.}
\label{fig:contact prob}
% \vspace{-0.5cm}
\end{figure}

We further study the efficacy of early stopping and the effect of applying smoothness loss to contact prediction as a substitute for our velocity smoothing. To this end, we use the \textbf{IEKF with NMN $\hat{\mathbf{c}}_t$} method to show how each part affects the contact prediction performance. The following $L_{sm}^c$ is the alternative smoothness loss for $\hat{\mathbf{c}}_t$, which is used instead of $L_{sm}$ for training.
\begin{equation} \label{loss function 2}
    L_{sm}^c = \frac{1}{N} (||_b\hat{\mathbf{c}}_t - _b\hat{\mathbf{c}}_{t-1}||^2 + \frac{1}{2} ||_b\hat{\mathbf{c}}_{t} - 2_b\hat{\mathbf{c}}_{t-1} + _b\hat{\mathbf{c}}_{t-2} ||^2 )
\end{equation}
The results in \tabref{early stopping} show that the performance deteriorates considerably, when $L_{sm}^c$ is used or as the training iteration extends. This phenomenon becomes clear upon viewing \figref{contact prob}. For the yellow line, which undergoes 1000 iterations without $L_{sm}^c$, the contact probability appears noisier than the red line. This noisy result is likely attributed to the NMN overfitting to simulation data, amplifying the sim-to-real gap. While actual contact states change quickly, the $L_{sm}^c$ impedes the rate of change in contact probability, leading to an uptick in false positive contact. Prolonged training accentuates this gradual shift in contact probability, thereby diminishing estimation accuracy.

\begin{table}[t]
\vspace{0.2cm}
\caption{\textbf{Ablation Study:} Impact of Smoothness Loss and Early Stopping on Flat Terrain Experiment}
\label{tab:early stopping}
\renewcommand{\arraystretch}{1.5}
\setlength{\tabcolsep}{5pt}
\resizebox{\columnwidth}{!}
{
\begin{tabular}{|c|c|c|c|c|c|c|c|c|}
\hline
\textbf{Terrain}           & \textbf{iteration} 
                           & \textbf{Method}   
                           & \textbf{\begin{tabular}[c]{@{}c@{}}ATE\\ (pos)\end{tabular}} 
                           & \textbf{\begin{tabular}[c]{@{}c@{}}ATE\\ (vel)\end{tabular}} 
                           & \textbf{\begin{tabular}[c]{@{}c@{}}ATE\\ (ori)\end{tabular}} 
                           & \textbf{\begin{tabular}[c]{@{}c@{}}RE\\ (pos)\end{tabular}} 
                           & \textbf{\begin{tabular}[c]{@{}c@{}}RE\\ (vel)\end{tabular}} 
                           & \textbf{\begin{tabular}[c]{@{}c@{}}RE\\ (ori)\end{tabular}} \\ \hline

\multirow{4}{*}{\textit{Flat}}     & \multirow{2}{*}{200}  & \textbf{w/o $L_{sm}^c$} & \textbf{0.3585} & \textbf{0.0791} & \textbf{0.0824} & \textbf{0.2418} & \textbf{0.1035} & \textbf{0.0397} \\ \cline{3-9}
                          &                       & w/ $L_{sm}^c$           & 0.6404 & 0.0803 & 0.0859 & 0.2700 & 0.1049 & 0.0398 \\ \cline{2-9}
                          & \multirow{2}{*}{1000} & w/o $L_{sm}^c$          & 1.3755 & 0.1184 & 0.0935 & 0.3872 & 0.1401 & 0.0404 \\ \cline{3-9}
                          &                       & w/ $L_{sm}^c$           & 1.1102 & 0.1037 & 0.1014 & 0.3625 & 0.1280 & 0.0405 \\ \hline 
                           
\end{tabular}
}
\end{table}

% \hspace{1.5cm}

\section{Conclusion And Future Work}
In this study, 
% we present a Neural Measurement Network (NMN) that can be integrated with IEKF to take advantage of model-based and data-driven state estimators.
we develop a state estimation framework that integrates a neural measurement network (NMN) with an invariant extended Kalman filter (IEKF) to make use of model-based and data-driven state estimators. 
NMN is composed of GRU-MLP and uses only proprioceptive sensors. The output of NMN is the contact probability and body linear velocity, which are used in the IEKF correction step. The proposed body linear velocity measurement model is suitable for the right-invariant observation form. Unlike previous studies, our approach does not use real-world data and relies solely on simulation data. Therefore, the sim-to-real gap needs to be reduced and to solve this, we analyze the efficacy of existing learning techniques, such as smoothness loss, domain randomization, and early stopping. 
% Therefore, the sim-to-real gap needs to be reduced, and to solve this, smoothness loss, avoiding domain randomization, and early stopping techniques were proposed. 

We collect data on various terrains with a quadrupedal robot Raibo and validate the estimator performance through Vicon data. These terrains include those with unstable and unclear contact conditions. Our method demonstrate its effectiveness by outperforming model-based state estimator. \deleted{Our research results may be helpful in subsequent research to reduce the sim-to-real gap.} Additionally, the linear velocity measurement model obtained through learning can be integrated into not only IEKF but also other types of estimators. Therefore, our research results will be helpful in the development of state estimators used in many fields beyond legged robots.
\section*{ACKNOWLEDGMENT}
This research was financially supported by the Institute of Civil Military Technology Cooperation funded by the Defense Acquisition Program Administration and Ministry of Trade, Industry and Energy of Korean government under grant No. UM22207RD2.

\bibliographystyle{IEEEtran}
\bibliography{IEEEabrv, ref}
% \input{modules/reference}

%%%%%%%%%%%%%%%%%%%%%%%%%%%%%%%%%%%%%%%%%%%%%%%%%%%%%%%%%%%%%%%%%%%%%%%%%%%%%%%%
\addtolength{\textheight}{-12cm}   % This command serves to balance the column lengths

%%%%%%%%%%%%%%%%%%%%%%%%%%%%%%%%%%%%%%%%%%%%%%%%%%%%%%%%%%%%%%%%%%%%%%%%%%%%%%%%

\end{document}